\documentclass[sigconf]{acmart}

\usepackage{multirow}
\usepackage{subcaption}

\AtBeginDocument{%
  }

\setcopyright{acmcopyright}
\copyrightyear{2018}
\acmYear{2018}
\acmDOI{XXXXXXX.XXXXXXX}

\acmConference[ICMR '23]{Make sure to enter the correct
  conference title from your rights confirmation emai}{June 12-15,
  2023}{Thessaloniki, Greece}
\acmPrice{15.00}
\acmISBN{978-1-4503-XXXX-X/18/06}

\begin{document}

\title{Multi-modal Fake News Detection on Social Media via Multi-grained Information Fusion}

\author{Yangming Zhou}
\email{ymzhou21@m.fudan.edu.cn}
\affiliation{%
  \institution{School of Computer Science\\ Fudan University}
  \streetaddress{2005 Songhu Rd}
  \city{Shanghai}
  \country{China}
}
\author{Yuzhou Yang}
\email{yzyang22@m.fudan.edu.cn}
\affiliation{%
  \institution{School of Computer Science\\ Fudan University}
  \streetaddress{2005 Songhu Rd}
  \city{Shanghai}
  \country{China}
}
\author{Qichao Ying}
\email{shinydotcom@163.com}
\affiliation{%
  \institution{School of Computer Science\\ Fudan University}
  \streetaddress{2005 Songhu Rd}
  \city{Shanghai}
  \country{China}
}

\author{Zhenxing Qian}
\authornote{Corresponding author. All the authors are also with the Key Laboratory of Culture and Tourism Intelligent Computing of Ministry of Culture and Tourism, Fudan University.}
\email{zxqian@fudan.edu.cn}
\affiliation{%
  \institution{School of Computer Science\\ Fudan University}
  \streetaddress{2005 Songhu Rd}
  \city{Shanghai}
  \country{China}
}

\author{Xinpeng Zhang}
\authornotemark[1]
\email{zhangxinpeng@fudan.edu.cn}
\affiliation{%
  \institution{School of Computer Science\\ Fudan University}
  \streetaddress{2005 Songhu Rd}
  \city{Shanghai}
  \country{China}
}

\renewcommand{\shortauthors}{Zhou et al.}

\begin{abstract}
The easy sharing of multimedia content on social media has caused a rapid dissemination of fake news, which threatens society's stability and security. Therefore, fake news detection has garnered extensive research interest in the field of social forensics.
Current methods primarily concentrate on the integration of textual and visual features but fail to effectively exploit multi-modal information at both fine-grained and coarse-grained levels. Furthermore, they suffer from an ambiguity problem due to a lack of correlation between modalities or a contradiction between the decisions made by each modality.
To overcome these challenges, we present a Multi-grained Multi-modal Fusion Network (MMFN) for fake news detection. 
Inspired by the multi-grained process of human assessment of news authenticity, we respectively employ two Transformer-based pre-trained models to encode token-level features from text and images.  The multi-modal module fuses fine-grained features, taking into account coarse-grained features encoded by the CLIP encoder. To address the ambiguity problem, we design uni-modal branches with similarity-based weighting to adaptively adjust the use of multi-modal features.
Experimental results demonstrate that the proposed framework outperforms state-of-the-art methods on three prevalent datasets.
\end{abstract}




\maketitle

\section{Introduction}
Online social networks, such as Twitter and Weibo, have largely replaced traditional forms of information communication represented by newspapers and magazines. Despite the convenience they offer to users, including the ability to seek out friends and share viewpoints, these networks have also facilitated the rapid and widespread dissemination of fake news~\cite{FND-Survey,ICASSP-1,MVAE}. 
 Compared to traditional written materials, online news posts are more susceptible to manipulation. Moreover, readers can be easily influenced by well-crafted fake news and may inadvertently further its spread.The consequences of fake news include the creation of panic, the misdirection of public opinion, and negative impacts on society. With hundreds of millions of social media users generating a vast amount of posts constantly, the use of traditional, inefficient manual review methods is not feasible for mitigating the threat of fake news. In response, many researchers in the field of computer science have recently focused on developing methods for detecting fake news~\cite{allein2021like,shu2020fakenewsnet,ma2016detecting,Entity-Enhanced, WWW}.

\begin{figure}[!t]
  \centering
  \includegraphics[width=0.95\linewidth]{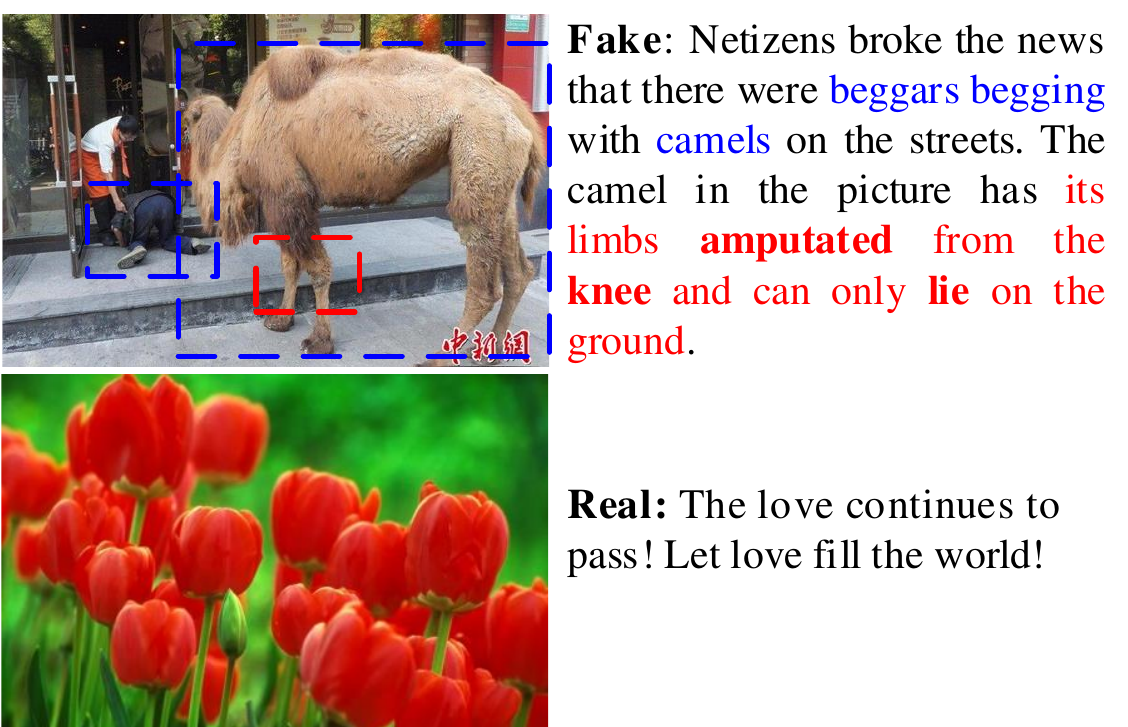}
  \caption{Examples of the two challenges faced by current works. Upper: multi-grained information is required to detect misinformation (the blue box represents matched elements and the red box represents mismatched elements). Lower: weak correlation between text and image may prevent users/models from finding cross-modal clues.
  }
  \label{intro}
\end{figure}

Early works on fake news detection primarily focused on analyzing either text-only or image-only content~\cite{WWW-4,Leveraging_6}. These works typically verify the logical and semantic coherence of the input and took into account trivial indicators, such as grammatical errors or traces of image manipulation.
Although uni-modal approaches are effective, modern news articles and posts often include multiple modalities that are inter-related, hence, these methods overlook such correlations.
For instance, a real image can be combined with total rumors and correct words can be used to describe a tampered image. 
Given this, multi-modal feature analysis is necessary to offer complementary benefits for fake news detection. 
In recent years, there have been many works that combine multi-modal features to detect anomalies in news~\cite{MVAE,EANN,WWW,ying2023boot}.
Despite the advancements made in the field, current approaches face two significant challenges. 
First, though many works come up with novel fusion methods, they merely fuse them at holistic post level, inevitably missing some detailed information; Or they only consider the matching between entities, tokens, or regions, neglecting global semantic correlation. This results in suboptimal utilization of information across different granularities.
Second, many works rely excessively on multi-modal fusion features, thus suffer from the ambiguity problem, i.e. inconsistency caused by inter-modal conflict or weak correlation. 


Figure~\ref{intro} illustrates two examples in the Weibo dataset that demonstrate the aforementioned two challenges. 
The upper image in Figure~\ref{intro} depicts the process of multi-grained fake news detection, wherein neither uni-modal feature of the text nor image is capable of verifying authenticity. 
First, readers of the post would pay attention to the objects in the image and entities in the text. 
For the examples given, people would first see the beggar and camel in the picture, and the words \emph{beggar}, \emph{begging}, \emph{camel}, \emph{amputation}, \emph{knee} and \emph{lie} in the text. At this time, other than matched elements (marked as blue region), they would find the three words \emph{amputation}, \emph{knee} and \emph{lie} do not match the image content (marked as red region). Subsequently, they will comprehend the semantic meaning of the sentence and image as a whole, perform an analysis to determine whether the two match, and finally, arrive at a conclusion regarding the authenticity of the news. 
Many existing works tend to ignore this point. 
The lower image of Figure~\ref{intro} is an example that illustrates the ambiguity and is also a common type of post on social media. The visual objects and the textual entities in the post have no matching relationship and are semantically unrelated. The user who posts this post may only be expressing their feelings. Manual review can easily determine that this is not a fake news and is not harmful. However, models that over-emphasize multi-modal fusion may misjudge it due to the mismatches in the multi-modal features. As a result, the post may not pass the automatic review process of social software and may not be published smoothly.
At this point, analyzing uni-modal content and emotions is sufficient to evaluate the credibility of the post, and multi-modal fusion features are not necessary and may even add noise to the classification task.
To address this issue, Chen et al.~\cite{WWW} proposed CAFE that first compresses the image and text and contrasts the news that has the correct image-text pairs to learn by minimizing the Kullback-Leibler (KL) divergence. Then, the corresponding cross-modal ambiguity score is used to reweight the multimodal features.
However, as only limited data is input into the network, their alignment of multi-modal features cannot be guaranteed. Furthermore, they process uni-modal features through variational autoencoders with non-shared weights, so we cannot determine if the KL divergence is reliable enough to measure cross-modal ambiguity. Thus, resolving the ambiguity problem remains a motivation for our research work.


To address the above-mentioned issues, this paper proposes the Multi-grained Multi-modal Fusion Network (MMFN).The MMFN approach integrates uni-modal features and a multi-grained multi-modal fused feature for more accurate fake news detection.
MMFN encodes text and image using pre-trained BERT~\cite{BERT} and Swin Transformer (Swin-T)~\cite{Swin} models respectively, which extract fine-grained information at the token level. 
The pre-trained CLIP~\cite{CLIP} model is utilized to encode coarse-grained features and capture semantic information at the post level, which is further used to address the ambiguity problem.
We propose a multi-grained multi-modal fusion module to perform granularity-level fusion on multi-modal features, in which fine-grained features are fed into co-attention Transformer (CT) blocks for generating aligned fine-grained multi-modal features, coarse-grained features are used for evaluating the cross-modal correlation. This module adjust the usage of multi-modal features dynamically, mitigating the ambiguity problem.
 

The contributions of this paper are mainly three-folded, namely:\begin{itemize}
\item We propose MMFN, which implements the idea of processing multi-modal features at different levels of granularity to form a comprehensive representation that reflects both the detailed and global aspects of the news.
\item We specifically design two uni-modal branches and adopt CLIP pre-trained model to evaluate cross-modal correlation, further address the problem brought by scenario with high cross-modal ambiguity.
\item We conduct comprehensive experiments on three famous datasets, where MMFN outperforms state-of-the-art fake news detection methods. What's more, ablation studies verify the effectiveness of granularity-level processing and multi-modal features adjustment. 
\end{itemize}


\section{Related Works}
\subsection{Fake News Detection}

Modalities refer to information perceived by different propagation channels that human communication is adapted to~\cite{Multimodal-Review}. Fake news detection methods could be categorized by utilized modal as uni-modal and multi-modal fake news detection, tons of work has been done on both methods based on deep learning in recent years. 


While uni-modal features like text content plays a key role in distinguishing fake news~\cite{MDFEND,M3FEND}, correlation and consistency of multi-modal features are also vital. 
Earlier works design sophisticated yet black-box attention mechanisms for multi-modal feature fusion~\cite{WWW-4,WWW-3,MVAE}.
Singhal et al. integrate pre-trained XLNet ResNet respectively for text and feature extraction. The features are then concatenated for the final binary classification~\cite{SpotFake}.
Singhal et al.~\cite{singhal2019spotfake} propose the Spotfake using VGG and BERT to extract features, which is then improved in  Spotfake+~\cite{SpotFake} for full-length article detection.
Wang et al.~\cite{EANN} propose an event advisory neural network, namely EANN, which tries to extract shared features among news of different events. 
Many other works~\cite{MVAE,WWW,ying2023boot} also propose better fuison of extracted features from different modalities before sending them into a classifier.

Compared with uni-modal methods, multi-modal methods not simply have extra image features, but have more potential information underneath multi-modal contents for mining. Some works try to excavate potential information from the datasets. 
Qi et al. ~\cite{Entity-Enhanced} claim former works neglect information that cannot be extracted by a pre-trained extractor, such as entities and texts within images. Thus, they manually extract these information as supplement of text content. 
Zhang et al. ~\cite{Dual-Emotion} design a novel dual emotion feature descriptor to measure the emotional gap between the publisher and comments and further verify that dual emotion is distinctive between fake and real news.
Allein et al.~\cite{allein2021like} propose DistilBert, which uses latent representations of news articles and user-generated content to guide model learning.
Wang et al.~\cite{wang2020fake} propose KMGCN that integrates textual, visual, and knowledge information into a unified framework to model semantic representations and improve accuracy
Further, Abdelnabi et al. ~\cite{CCN-CVPR2022} use online search engine to gather relevant Web evidence and propose Consistency-Checking Network to mimic human reasoning process.

Except from mining information, planning better multi-modal representations’ interaction is vital for better detection performance.
SAFE~\cite{zhou2020mathsf} calculates the relevance between news textual and visual information.
MCAN ~\cite{wu2021multimodal} stacks multiple co-attention layers to fuse the multi-modal features.
Qian et al. ~\cite{HMCAN} use BERT to produce hierarchical semantics of text, and use ResNet to produce regional image representations. Then different levels of semantics are fed into co-attention layers with regional image features to achieve hierarchical multi-modal feature fusion. 
FND-CLIP ~\cite{zhou2023clip} uses CLIP-exctracting features as multimodal represention and designs a modal-wise attention to aggregat features.

\subsection{Pre-trained Model}


Vaswani et al. ~\cite{vaswani2017attention} propose Transformer, which quickly become the state-of-the-art method in many NLP tasks. Subsequently,
Devlin et al. ~\cite{BERT} propose BERT, a large Transformer-based model pre-trained on large corpora.
Transformer-based image processing methods also gained merits in their fields. However, a simple application of self-attention to images would require quadratic cost in the number of pixels, causing a problem with input sizes of realistic images. 
Dosovitskiy et al. ~\cite{dosovitskiy2020image} directly apply a Transformer architecture on non-overlapping medium-sized image patches for image classification, namely ViT. On top of that, Liu et al. ~\cite{Swin} propose Swin-T combining convolution layer with proposed Swin-T layers, achieving linear increase in complexity with image size. Swin-T focuses on general-purpose performance rather than specifically on classification like ViT, yet still achieves the best speed-accuracy trade-off among other methods on image classification. 

In the past decade, multi-modal machine learning has received considerable attention in the research community~\cite{Multimodal-Review}.
Neural architectures are employed in tasks that go beyond single modalities, for example, Visual Question Answering (VQA)~\cite{VQA}, Visual Commonsense Reasoning (VCR)~\cite{VCR}, etc. 
In these tasks, priors and features from different modalities are required and algorithms or deep networks cannot be effective when provided with only a single modality.
Several generic technologies are developed for learning joint representations of image content and natural language.
For example, CLIP~\cite{CLIP} is a multi-modal model that combines knowledge of language concepts with semantic knowledge of images. Benefiting from the contrastive learning paradigm, the textual and visual features extracted by CLIP can be considered to be aligned in the same semantic space, which can reflect the correlation between the textual content and visual content of news.
The CLIP-based Multi-modal learning has been used in a lot of downstream tasks~\cite{Hair-CLIP,CLIP-Art}.
Other multi-modal models, such as Glide~\cite{Glide} and VilBERT~\cite{vilbert}, have also been utilized for tasks such as text-to-image generation and cross-modal representation learning.
\begin{figure*}[!t]
  \centering
  \includegraphics[width=0.80\linewidth]{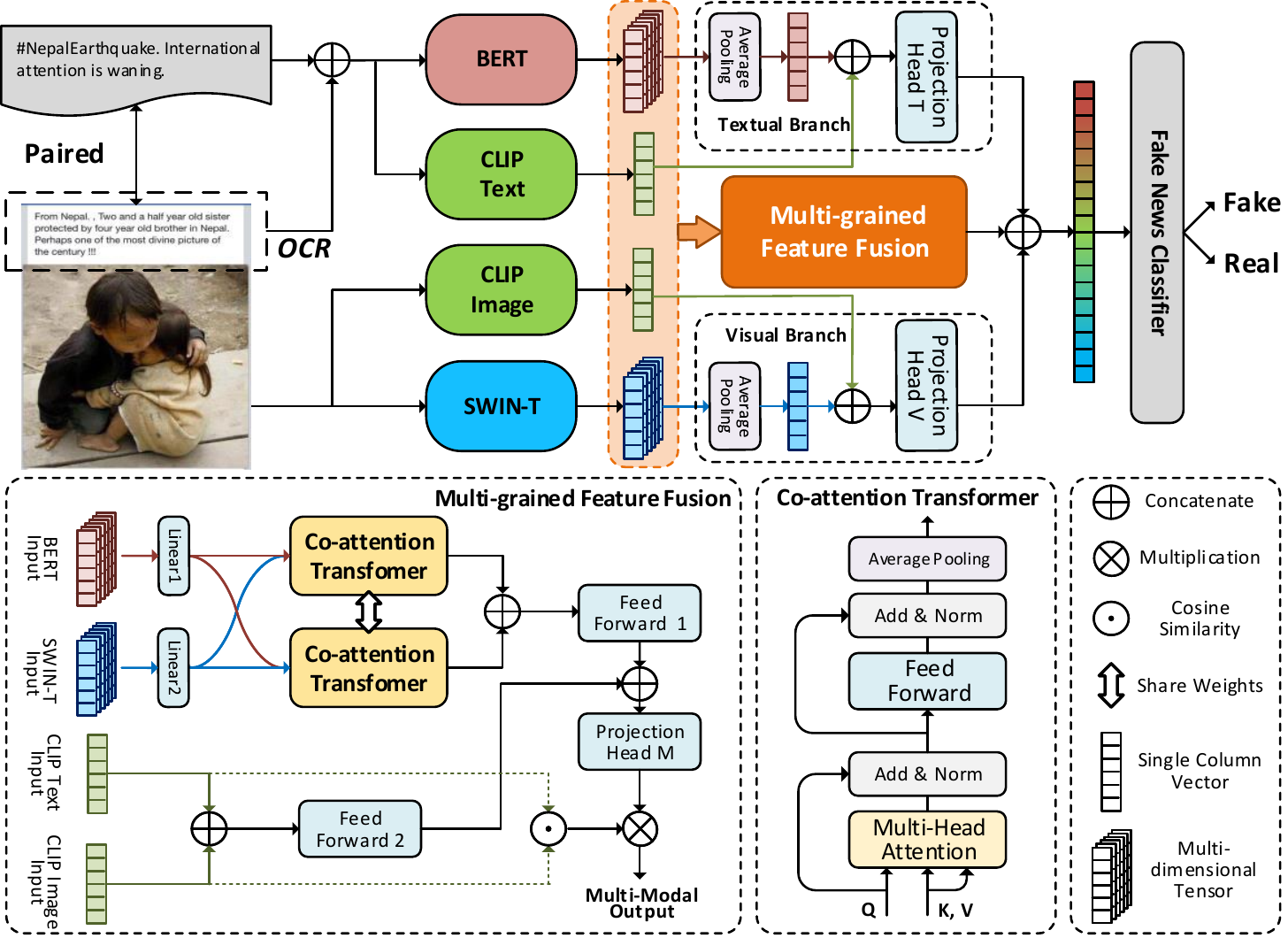}
    \caption{The architecture of the MMFN. BERT, Swin-T, and CLIP are used to encoded the features of different modalities of multi-modal news. Encoded features of different granularities are fused through CT and projection heads. CLIP similarity score is calculated to weight different modal features adaptively for the classifier to classify fake news.}
  \label{architecture}
\end{figure*}

\section{METHOD}
In this paper, we propose MMFN to improve the accuracy of multi-modal fake news detection via multi-grained multi-modal fusion. The network design of MMFN is shown in Figure~\ref{architecture}, which consists of multi-modal feature encoder,  multi-grained feature fusion, uni-modal branches and modality weighting via CLIP similarity, and fake news classifier.
\subsection{Multi-modal Feature Encoder}
\noindent\textbf {Textual Feature Encoding via BERT.}
BERT~\cite{BERT} is a popular pre-trained language model built on Transformer that is trained using unsupervised learning on a large corpus and has achieved excellent results in many NLP downstream tasks. Therefore, we use a BERT model to encode features from $\mathbf{T}$. The textual content of a news post, which is the concatenation of text and optical character recognition (OCR) extraction from an image, is a sequential list of words denoted as ${{\mathbf{T}}}=\left[ t_{1},t_{2},\ldots ,t_{n_w} \right]$, where ${n_w}$ is the number of words. After applying BERT to $\mathbf{T}$, the encoded textual feature $\mathbf{T}^b =\left[ t^b_{1},t^b_{2},\ldots ,t^b_{n_w} \right] $ is obtained, where ${t^b_{i}\in \mathbb{R}^{d_b}}$ is the output for the last hidden state of the ${i-}$th token in the text embedding and ${d_b}$ is the dimension of the word embedding.

\noindent\textbf {Visual Feature Encoding via Swin-T.}  
 Swin-T ~\cite{Swin} is a vision Transformer that produces a hierarchical feature representation based on shifted window self-attention and achieves state-of-the-art performance on many CV tasks such as object detection and semantic segmentation. In this paper, we introduce Swin-T to the task of fake news detection. Given the visual content $\mathbf{V}\in \mathbb{R}^{w \times h}$, Swin-T transforms it to a sequence embeddings $\mathbf{V^s} =\left[ v^s_{1},v^s_{2},\ldots ,v^s_{n_p} \right]$, where $w$ and $h$ are the width and height of the image, ${v^s_{i}\in \mathbb{R}^{d_s}}$ is the hidden-states at the output of the last layer of the model corresponding to the $i-$th window of the input, $n_p$ is the number of patch in Swin-T, and ${d_s}$ is the hidden size of the visual embedding.

\noindent\textbf {Multi-modal Feature Encoding via CLIP.} 
CLIP is pre-trained on a massive and diverse dataset, and can embed texts and images into a uniform mathematical space, making it naturally propitious to calculate the cross-modal correlation. Besides, experiments have shown the model’s robustness in dealing with distribution shift, making it suitable for fake news detection even in zero-shot scenarios~\cite{CLIP}. Many works~\cite{ViLD,li2022language} have shown CLIP's great ability in generalizing to unknown fields.
 Therefore, we use CLIP multi-modal features to enrich the global correlation information of textual and visual features. Given the multi-modal news $\mathbf{X}=\{\mathbf{T},\mathbf{V}\}$, we denote the CLIP-encoded features as $\mathbf{X}^c=\left[{t^c},{v^c}\right]$, where ${t^c, v^c\in \mathbb{R}^{d_c}}$ are two vectors of length ${d_c}$.
\subsection{Multi-grained Feature Fusion}
\noindent\textbf {Fine-Grained Fusion via Transformer.} 
Since BERT and Swin-T are not multi-modal models, there is a large gap between the features they extract, which cannot directly realize information interaction. To effectively fuse the textual and visual features of posts, we use a CT~\cite{vilbert} to achieve information inter-modal complementarity. As shown in Figure~\ref{architecture}, CT consists of a multi-headed attention network and a feed forward neural network, both followed by a residual connection and a layer normalization. 

We denote different modal inputs as ${I_1}$ and ${I_2}$, respectively. In CT, ${I_1}$ is used as queries $Q$, and ${I_2}$ is used as keys $K$ and values $V$. CT computes the co-attention matrix of each head as:
\begin{equation}
h_i={Softmax}(\frac{(I_1 W_i^q)(I_2 W_i^k)^T}{d^h})I_2  W_i^v \text {, }
\end{equation}
where the projection matrices $W_i^q, W_i^k, W_i^v\in \mathbb{R}^{d_m \times d_h}$, $d_h=d_m/m$, and $d_m$ is the dimension of the CT model, while $m$ is the number of heads. 

The multi-head attention is the concatenation of all co-attention matrices following a projection matrix:
\begin{equation}
H=(h_1;h_2;\cdots;h_m)W^o \text {, }
\end{equation}
where symbol $;$ represents the concatenate operation and $W^o\in \mathbb{R}^{d_m\times d_m}$.

After $H$ and ${I_1}$ pass through the FFN with two layer normalization, an attention-based multi-modal representation $H^{\prime}$ is obtained:
\begin{equation}
H^{\prime} = \operatorname{Norm} ({I_1} + \operatorname{FFN}(\operatorname{Norm}({I_1} + H{I_1}))) \text {. }
\end{equation}

Finally, the multi-modal representation $H^{\prime}$ is average pooled to a feature vector $F$ as the output of CT:
\begin{equation}
F=\operatorname{CT}(I_1, I_2) = \overline{H^{\prime}} \text {. }
\end{equation}

In our model, after the inputs BERT and Swin-T features $\mathbf{T}^b, \mathbf{V}^s$ are mapped to the same dimension ${\mathbb{R}^{d_m}}$ through linear layers, they are input into a shared weighted CT in different front and rear order as ${I_1}$ and ${I_2}$ to obtain the output features, a visual attention weighted textual feature ${F^{vt}}$ and a textual attention weighted visual feature ${F^{tv}}$, respectively:
\begin{equation}
\left\{\begin{array}{l}
{F^{vt}}=\operatorname{CT}((\mathbf{T}^bW^t), (\mathbf{V}^sW^v)) \\
{F^{tv}}=\operatorname{CT}((\mathbf{V}^sW^v), (\mathbf{T}^bW^t)) 
\end{array}\right. \text {, }
\end{equation}
where ${W^t\in \mathbb{R}^{d_b \times d_m} }$ and ${W^v\in \mathbb{R}^{d_s \times d_m} }$. 

MCAN~\cite{wu2021multimodal} and HMCAN~\cite{HMCAN} also use Transformer to fuse features. There are structural and functional differences between the CT in MMFN and the Transformers in HMCAN and MCAN.
MCAN takes the pooled feature vectors as the input to the stacked co-attention blocks, which outputs a coarse-grained modality correlation vector; while the attention structure of MMFN is calculated at the token level and outputs fine-grained fused features.
The Transformer in HMCAN consists of a single-head self-attention to enhance intra-modal information and another one to capture inter-modal information; considering BERT and Swin-T already use self-attention to achieve intra-modal interaction, to avoid impacting pre-training model and alleviate the overfitting problem, CT of MMFN only includes a multi-head co-attention to achieve inter-modal interaction. In addition, the Transformers in HMCAN do not share weights, while Transformers of MMFN share weights to make CT have the function of modal alignment.

\noindent\textbf {Coarse-grained Fusion via CLIP and Multi-modal Representation Generation.} 
We fuse the CLIP-encoded features as robust coarse-grained features that reflect the global semantic correlation information. It can be assumed that the outputs ${t^c}$ and ${v^c}$ of the CLIP encoders have effectively eliminated the inter-modality gap through contrastive learning. This allows subsequent network learning to effectively utilize the information from different modalities.

Specifically, ${t^c}$ and ${v^c}$ are concatenated and fed into a feed forward neural network with a linear layer, a batch norm layer, and a ${ReLU}$ activation function. Additionally, ${F^{vt}}$ and ${F^{tv}}$ are also concatenated and fed into another feed forward neural network with the same architecture. Thus, after two feed forward networks, we have the fused multi-modal fine-grained feature ${M^f}$ and the coarse-grained feature ${M^c}$: 
\begin{equation}
\left\{\begin{array}{l}
{{M}^{f}}=\operatorname{FFN_1}(F^{vt};F^{tv}) \\
{{M}^{c}}=\operatorname{FFN_2}(t^c;v^c)
\end{array}\right. \text {. }
\end{equation}

Finally, the multi-modal features ${M^f}$ and ${M^c}$ are concatenated and fed into a projection head ${\Phi_M}$ to generate the multi-modal representation. 
Each element in the output vector will be multiplied by a similarity score, which will be introduced in the next paragraph. 
The generated multi-modal representation is denoted as ${F^m}$:
\begin{equation}
F^m = \emph{similarity}\cdot{\Phi_M{(M^f;M^c)}} \text {. }
\end{equation}
\subsection {Uni-modal Branches and Modality Weighting via CLIP Similarity} 
Multi-modal fused features generally reflect the correlation information between the two modalities, which is easily affected by ambiguity. To solve the problem that the feature representation ability of multi-modal fusion decreases when the modal is of high ambiguity, we designed a uni-modal textual branch and a uni-modal visual branch respectively.

For the textual branch, we pool the BERT features into a feature vector on the dimension of token level, concatenate it with the CLIP-text feature vector, then pass it through a projection head consisting of two fully connected networks with ${ReLU}$ activation functions to get the uni-modal textual representation; similarly, for the visual branch, we concatenate the pooled Swin-T features with the CLIP-image feature, and obtain the uni-modal visual through a projection head mapping with the same structure but different parameters as the textual branch's:
\begin{equation}
\left\{\begin{array}{l}
{{F}^{t}}=\Phi_T(\overline{\mathbf{T}^b};t^c) \\
{{F}^{v}}=\Phi_V(\overline{\mathbf{V}^s};v^c)
\end{array}\right. \text {, }
\end{equation}
where $\Phi_T$ and $\Phi_V$ are the projection heads.

If we directly send the uni-modal branch representations to the classifier for making the decision, the classifier may be more inclined to use the multi-modality representation with deeper network to fit the results, while the uni-modal branch could interfere with the decision and cause more serious ambiguity problems. To overcome such limitations, inspired by CAFE, we use the CLIP cosine similarity as a coefficient weighting the multi-modal feature to guide the classifier's learning process. The cosine similarity is calculated as follows:
\begin{equation}
\emph{similarity}=\frac{{{t}^{c}}\cdot {{({{v}^{c}})}^{T}}}{\left\| {{t}^{c}} \right\|\left\| {{v}^{c}} \right\|} \text {. }
\end{equation}
\subsection{Fake News Classifier}
After obtaining the fused multi-modal representation, uni-modal text representation, and uni-modal visual representation, we concatenate them as the input to a classifier and get the output $\hat{y}$ representing the probability of a news being fake:
\begin{equation}
\hat{y}= \operatorname{FNC}(F^m;F^t;F^v) \text {, }
\end{equation}
where $\operatorname{FNC}(\cdot)$ is the fake new classifier consisting of a two-layer fully connected network with ${ReLu}$ activation function.

The objective function is to minimize the cross-entropy loss to correctly predict the real and fake news.
\begin{equation}
\mathcal{L}=y \log \left( \hat{y}\right)+\left(1- y\right) \log \left(1- \hat{y}\right) \text {, }
\end{equation}
where $y$ is the ground truth label.
\begin{table*}[!t]
\renewcommand\arraystretch {0.9}
  \caption{\noindent\textbf{Performance comparison between MMFN and state-of-the-art methods on Weibo, Twitter, and Gossipcop datasets. The best performance is highlighted in bold.} }
  \label{comparison}
  \setlength{\tabcolsep}{3mm}{
\begin{tabular}{ccccccccc}
\toprule
\multirow{2}{*}   & \multirow{2}{*}{Method} &\multirow{2}{*}{Accuracy} &\multicolumn{3}{c}{Fake News}  & \multicolumn{3}{c}{Real News}\\
   \cline{4-9}&       &     & Precision   & Recall  & F1-score    & Precision    & Recall      & F1-score             \\
\midrule
\multirow{6}{*}{Weibo} 
& Spotfake~\cite{singhal2019spotfake}    & 0.892    & 0.902  & \textbf{0.964}   & \textbf{0.932}   & 0.847    & 0.656    & 0.739   \\
& CAFE~\cite{WWW}    & 0.840     & 0.855  & 0.830  & 0.842  & 0.825   & 0.851  & 0.837    \\
 & MCAN~\cite{wu2021multimodal}   & 0.899    & 0.913   & 0.889   & 0.901   & 0.884   & 0.909   & 0.897 \\ 
&LIIMR~\cite{singhal2022leveraging}    & 0.900    & 0.882   & 0.823   & 0.847    & 0.908    & \textbf{0.941}   &\textbf{0.925} \\
 & HMCAN~\cite{HMCAN}      & 0.885    & 0.920   & 0.845   & 0.881  & 0.856   & 0.926   & 0.890   \\
& \textbf{MMFN}   &  \textbf{0.923}  & \textbf{0.921} & 0.926   &0.924 & \textbf{0.924}  & 0.920   & 0.922\\
\midrule
\multirow{6}{*}{Twitter} 
& Spotfake~\cite{singhal2019spotfake}   & 0.777   & 0.751  & \textbf{0.900}  & 0.820  & 0.832   & 0.606    & 0.701   \\
& CAFE~\cite{WWW}    & 0.806     & 0.807  & 0.799  &  0.803  & 0.805   &  0.813  & 0.809    \\
 & MCAN~\cite{wu2021multimodal}      & 0.809    & 0.889   & 0.765   & 0.822   & 0.732   & 0.871   & 0.795   \\
&LIIMR~\cite{singhal2022leveraging}    & 0.831   & 0.836  & 0.832   & 0.830    &0.825   & 0.830   & 0.827 \\
 & HMCAN~\cite{HMCAN}      & 0.897    & \textbf{0.971}  & 0.801    & 0.878  & 0.853   & 0.979   &0.912   \\
& \textbf{MMFN}  & \textbf{0.935}  & 0.960 &  0.856   & \textbf{0.905} & \textbf{0.924} & \textbf{0.980}  & \textbf{0.951}\\
\midrule
\multirow{5}{*}{Gossipcop} 
& SAFE~\cite{zhou2020mathsf}  & 0.838  & 0.758 &0.558 & 0.643   & 0.857  & 0.937  & 0.895  \\
& Spotfake+~\cite{SpotFake}   & 0.858   & 0.732 & 0.372  & 0.494  &  0.866   &  0.962   & 0.914   \\
& DistilBert~\cite{allein2021like}  & 0.857  &  \textbf{0.805}  & 0.527  & 0.637  & 0.866& 0.960& 0.911\\
& CAFE~\cite{WWW} & 0.867 & 0.732  & 0.490  & 0.587   & 0.887& 0.957  & 0.921 \\

& \textbf{MMFN}   & \textbf{0.894}  & 0.799 & \textbf{0.598}  & \textbf{0.684} & \textbf{0.910} &  \textbf{0.964}   & \textbf{0.936} \\
    \bottomrule
  \end{tabular}}
\end{table*}

\section{EXPERIMENTS}
\subsection{Experimental Setup}
\noindent\textbf {Dataset.} We use three public real-world datasets collected from social media, namely, Weibo~\cite{jin2017multimodal}, Twitter~\cite{Twitter}, and Gossipcop~\cite{shu2020fakenewsnet}. 

Weibo is a widely used Chinese dataset in fake news detection. The real news was collected from Xinhua News Agency, an authoritative news source of China. In experiments, the uni-modal news posts with no image or no text description were filtered out. The training set contains 3, 783 real news and 3, 675 fake news, and the test set contains 1,685 news. 

The Twitter dataset was released for MediaEval Verifying Multimedia Use task~\cite{Twitter} and is also a well-known multi-modal dataset for fake news detection. In experiments, following existing works we filter the tweets with videos attached and the non-English tweets. After filtering, the training set contains 4, 031 real news and 5, 139 fake news, and the test set contains 1, 406 posts.

Gossipcop dataset is a English full-length article news dataset collected from the entertainment domain of FakeNewsNet~\cite{shu2020fakenewsnet} repository. 
Gossipcop contains 10, 010 training news, including 7, 974 real news and 2, 036 fake news. The test set has 2, 285 real news and 545 fake news.

\noindent\textbf {Implementation Details.}
In our experiments, we set ${d_m = 512}$ and ${m = 8}$ for the CT. 
The textual embedding dimension of BERT is set to ${d_b = 768}$, and we use the “bert-base-chinese” model for Chinese data and the “bert-base-uncased” model for English data. The input text length is set to 300 words, i.e., ${n_w = 300}$. For the Swin-T, we use the “swin-base-patch4-window7-224” model to encode visual features and set the input image size to 224 × 224. The number of patches in Swin-T is ${n_p = 49}$ and the dimension of the visual embedding is ${d_s= 1024}$. The input image size for CLIP is also set to to 224 × 224. Science CLIP has not pre-trained Chinese text model, we use Google Translation API~\cite{johnson2012google} to translate Chinese texts to English. We use a summary generation model~\cite{raffel2019exploring} to generate summary statements for texts longer than 50 words to meet the input size requirements of CLIP.  The used pre-trained CLIP model is “ViT-B/32” with feature dimension of ${d_c=512}$. We fine-tune BERT and Swin-T during the training stage, while freezing the parameters of CLIP due to its difficulty in training on small datasets.  Similar to MCAN ~\cite{wu2021multimodal}, we freeze the BERT model on the Twitter dataset to alleviate overfitting. The feed forward network 1 and the feed forward network 2 have hidden size of 256, and the projection heads had hidden units of  256 and 16, respectively. The hidden sizes of the two fully conneced layers in the classifier are 48 and 2, respectively. The batch size is set to 16. 
The dataset was pre-processed to remove all invalid messages after "@", "{\#}", and "http:". The Adam optimizer~\cite{kingma2014adam} was utilized with the default parameters. Aside from fine-tuning BERT and Swin Transformer using a learning rate of $1\times {{10}^{-5}}$, the learning rate for the network was set to $1\times {{10}^{-3}}$. The model is trained for 100 epochs with early stopping to prevent over-fitting.

\begin{table*}[!ht]
\renewcommand\arraystretch {0.9}
    \caption{Ablation study on modalities, granularities, and architecture designs of MMFN on Weibo, Twitter, and Gossipcop datasets. The best performance is highlighted in bold.}
\label{modal}
  \setlength{\tabcolsep}{3mm}{
\begin{tabular}{ccccccccc}
\toprule
\multirow{2}{*}   & \multirow{2}{*}{Method} &\multirow{2}{*}{Accuracy} &\multicolumn{3}{c}{Fake News}  & \multicolumn{3}{c}{Real News}\\
  \cline{4-9}  &     &  & Precision    & Recall    & F1-score   & Precision    & Recall    & F1-score   \\
\midrule
\multirow {8}{*}{Weibo}  
& MMFN w/o T  & 0.825  & 0.846  &  0.815    & 0.830      & 0.803 &0.836   & 0.819   \\
& MMFN w/o V  & 0.901  &0.859 &  0.940     & 0.898   &0.944  & 0.868     & 0.904   \\
& MMFN w/o F  & 0.818    & 0.793 & 0.839   & 0.815   & 0.844  & 0.800    & 0.821\\
& MMFN w/o C  & 0.909    & 0.914 & 0.907  & 0.911   & 0.904  & 0.912    & 0.908   \\
& MMFN w/o CT  & 0.905 & 0.849 & \textbf{0.959 }  & 0.900   & \textbf{0.963}  & 0.861    & 0.909   \\
& MMFN w/o U  & 0.912 & \textbf{0.939} & 0.891   & 0.915  & 0.884  &  \textbf{0.934}    & 0.908   \\
& MMFN w/o W  & 0.906 & 0.921 & 0.895   & 0.908   & 0.890  & 0.917    & 0.903   \\
& \textbf{MMFN}   & \textbf{0.923}   & 0.921 & 0.926   & \textbf{0.924} & 0.924  & 0.920   & \textbf{0.922}\\
\midrule
\multirow {8}{*}{Twitter} 
& MMFN w/o T  & 0.933  &0.963  &  0.850 & 0.903 & 0.919 &0.981   & 0.949   \\
& MMFN w/o V  & 0.739  &0.691& 0.580 & 0.630   &0.762  & 0.839     & 0.799   \\
& MMFN w/o F & 0.866    & \textbf{0.993} & 0.708   & 0.826   &0.805  & \textbf{0.996}    & 0.891  \\
& MMFN w/o C  &0.882  &0.817 &0.817 &0.817 &0.913 &0.913 &0.913  \\
& MMFN w/o CT  &  0.900    & 0.965 & \textbf{0.885}   & \textbf{0.923}   & 0.793  & 0.932    & 0.857   \\
& MMFN w/o U  & 0.918 & 0.929 &  0.835   & 0.880   & 0.913  & 0.965    & 0.938   \\
& MMFN w/o W  & 0.915 & 0.960 & 0.810   & 0.879   & 0.893  & 0.979    & 0.934   \\
& \textbf{MMFN}  & \textbf{0.935}  & 0.960 &  0.856   &  0.905 & \textbf{0.924} & 0.980  & \textbf{0.951 }\\
\midrule
\multirow {8}{*}{Gossipcop} 
& MMFN w/o T  & 0.836  &0.702  &  0.255 & 0.374 & 0.846 &0.974   & 0.905   \\
& MMFN w/o V  & 0.888  &0.818& 0.536 & 0.648   &0.898 & 0.972    & 0.933   \\
& MMFN w/o F & 0.862    & 0.691 & 0.517   & 0.592  &0.891  & 0.945    & 0.917 \\
& MMFN w/o C  &0.885  &\textbf{0.844} &0.495 &0.624 &0.890 &\textbf{0.978} &0.932  \\
& MMFN w/o CT  &  0.888    & 0.756 & \textbf{ 0.615}   & 0.678   & \textbf{0.912}  & 0.953    & 0.932   \\
& MMFN w/o U  & 0.885 & 0.799 &  0.539   & 0.644  & 0.898  & 0.965    & 0.932   \\
& MMFN w/o W  & 0.889 & 0.764 &0.611   & 0.679   & 0.911  & 0.968    & 0.933   \\
& \textbf{MMFN} & \textbf{0.894}  & 0.799 & 0.598 & \textbf{0.684} & 0.910 &  0.964   & \textbf{0.936} \\
\bottomrule
\end{tabular}}
\end{table*}
\subsection{Performance Comparison}
We compare the performance of MMFN with other state-of-the-art methods and the comparison results are presented in Table~\ref{comparison}. The evaluation metrics are accuracy, precision, recall, and F1-score, which are commonly used to measure the performance of a binary classification problem.
As shown in Table~\ref{comparison}, MMFN outperforms the other methods across all three datasets in terms of Accuracy. Specifically, MMFN achieves the highest accuracy of \textbf{92.3\%}, \textbf{93.5\%}, and \textbf{89.4\%}, respectively, on the three real-world datasets, surpassing the state-of-the-art method by \textbf{2.3\%}, \textbf{3.8\%}, and \textbf{2.7\%}. In terms of precision, recall, and F1 score, MMFN ranks either first or second on nearly all tests, showcasing its effectiveness.

The compared methods, SAFE, Spotfake, DistilBert and Spotfake+ have limitations in their methods for fake news detection. SAFE learns similarity between text and visuals, but may misclassify real posts with weak correlation as fake news due to ignoring ambiguity.
DistilBert guides the detection by checking the effect of user-related constraints on article latent space, yet it neglects visual information of news thus lead to less competitive performance.
Spotfake and Spotfake+ simply concatenate textual and visual representations without adequate cross-modal interaction and fusion, leading to suboptimal performance. CAFE defines and utilizes cross-modal ambiguity to alleviate the problem of disagreement between different modalities. It achieves better experimental results than Spotfake, Spotfake+, and SAFE on the Twitter and Gossipcop datasets. 

LIIMR achieves improved results on the Weibo dataset due to its ability to capture fine-grained salient image and text features. The decent experimental results of MCAN and HMCAN prove the effectiveness of the Transformer-based multi-modal fusion network. However, these approaches concentrate only on fine-grained feature mining, neglecting the coarse-grained information that provides insight into global semantics.

The superiority of MMFN over the other methods can be attributed to three factors.  1) The Swin-T component is capable of extracting fine-grained features that complement the features generated by the BERT encoder. Additionally, the pre-trained CLIP encoder is capable of generating coarse-grained text and image features that possess rich semantic information within a shared semantic space. This allows for complementary feature representation at both fine and coarse granularities.
2) The CT component enables inter-modal interaction at the token level, thus facilitating the fine-grained fusion of multiple modalities.
3) The utilization of uni-modal branches with CLIP-based weighting effectively mitigates the issue of ambiguity.

\subsection{Ablation Studies}
We evaluate the impact of the key components in MMFN on its performance by conducting experiments with various and partial configurations of the model.  For each experiment, we remove a different component and retrain the model from scratch.
The compared variants of MMFN are implemented as follows:

1) MMFN w/o T.  The text-related modules are removed and only the uni-modal visual features encoded by Swin-T and CLIP image coder are used.

2) MMFN w/o V. The visual-related modules are removed, and only the uni-modal textual features encoded by BERT and CLIP text coder are retained.

3) MMFN w/o F. The BERT-related and Swin-T-related modules are removed, and the fine-grained features are not utilized. Instead, the CLIP-coded text and image features are directly concatenated into a multi-modal representation, and the two CLIP features are used as two separate uni-modal representations.

4) MMFN w/o C. The CLIP-related modules are removed, and the coarse-grained features are not employed. The classification process is performed using only fine-grained features.

5) MMFN w/o CT. The CT module is removed, and the features encoded from BERT and Swin-T are directly concatenated.

6) MMFN w/o U. The textual and visual uni-modal branches are removed, and only the multi-modal fused representation is used for classification.

7) MMFN w/o W. The CLIP-weighting modules are removed, and the multi-modal fused feature is not weighted.

\noindent\textbf{Contributions from Different Modalities. }
To assess the contributions of different modalities to the overall performance of MMFN model, we compare the results of MMFN w/o T and MMFN w/o V with the completed MMFN model. The results indicate that the performance of MMFN decreases when either the visual or textual modality is absent. This suggests that both modalities are crucial for the final performance of the model.

Additionally, we found that the performance of MMFN w/o T is poor on the Weibo and Gossipcop datasets, but second-best on the Twitter dataset. On the other hand, MMFN w/o V performs poorly on the Twitter dataset, but outperforms the MMFN w/o T and some multi-modal ablated models on Weibo and Gossipcop. These observations indicate that: 1) The distribution of news on different datasets and different social media platforms is diverse. For instance, the text of Weibo news contains rich clues for fake news detection, while Twitter users tend to express information through images. Meanwhile, the main information in long news on Gossipcop is located in the text, while images play a minor role. This implies that uni-modal models may have difficulties adapting to different social environments.
2) The uni-modal models outperform several of the ablated multi-modal models, suggesting that multi-modal learning is effective only when the cue information from different modalities is adequately mined and integrated using an effective mechanism.

\noindent\textbf{Influence of Different Granularities. }
To analyze the effect of different granularities on the performance of the model, we compare the results of MMFN w/o C and MMFN w/o F with the completed MMFN model. The results show that both coarse-grained and fine-grained features contribute to the performance of the model. The performance of the model decreases more when the fine-grained features are removed than when the coarse-grained features are removed. This suggests that fine-grained features play a more important role in the final performance of the model. The suggestion follows our intuition, and mining more fine-grained features to detect fake news is the direction of the recent works like LIIMR and HMCAN, etc. However, the results of the ablation study shows that the coarse-grained features is useful to assist the fine-grained features to improve the detection ability.

\noindent\textbf{Effectiveness of Each Component.}
To evaluate the effectiveness of each component in the MMFN model, we compare the results of the completed MMFN model with the ablated models MMFN w/o CT, MMFN w/o U, MMFN w/o W. 
As can be seen in Table~\ref{modal}, the following observations are made:
1)The integration of visual and textual modalities through the CT module enhances the representation capability and fine-grained feature fusion, which is reflected in the improved performance of the complete model compared to the MMFN w/o CT. 
2) The Completed MMFN model outperforms MMFN w/o U, demonstrating the effectiveness of the uni-modal branch designed to handle ambiguity in improving the performance.
3) The results also show that both the complete MMFN model and MMFN w/o U outperform MMFN w/o W on the Weibo and Twitter datasets, indicating that weighting the fusion features helps alleviate ambiguity. On the other hand, simply adding uni-modal branches without weighting may harm performance.
Overall, the results suggest that the full combination of components is necessary for achieving the best performance of the MMFN model.

\begin{figure}[htbp]
    \centering
    \begin{subfigure}{0.22\textwidth}
      \centering
        \includegraphics[height=0.92\linewidth]{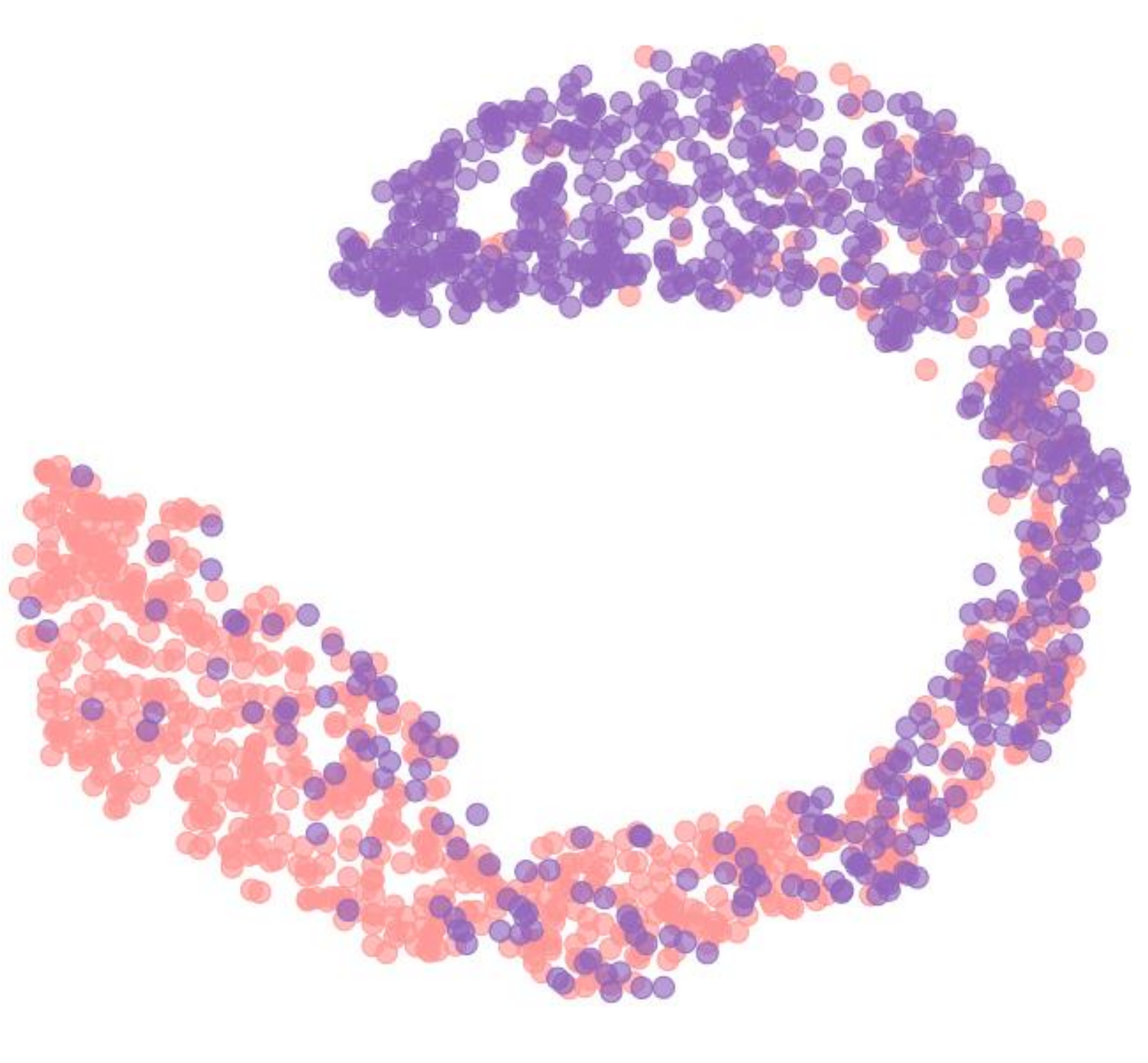}
        \caption{Visual representations}
        \label{image}
    \end{subfigure}
    \centering
    \begin{subfigure}{0.22\textwidth}
      \centering
        \includegraphics[height=0.92\linewidth]{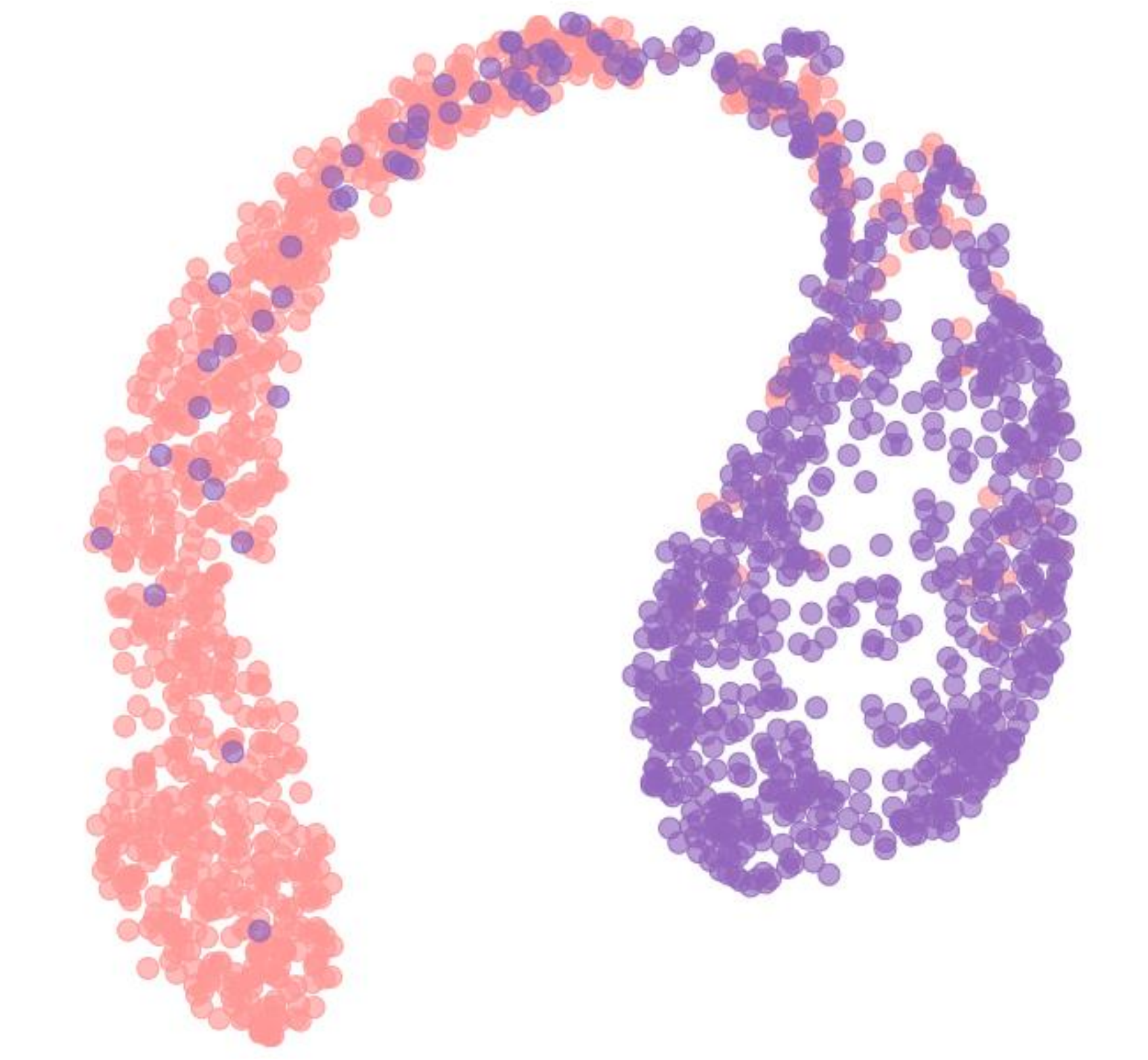}
        \caption{Textual representations}
        \label{text}
    \end{subfigure}
  \centering
    \begin{subfigure}{0.22\textwidth}
      \centering
        \includegraphics[height=0.92\linewidth]{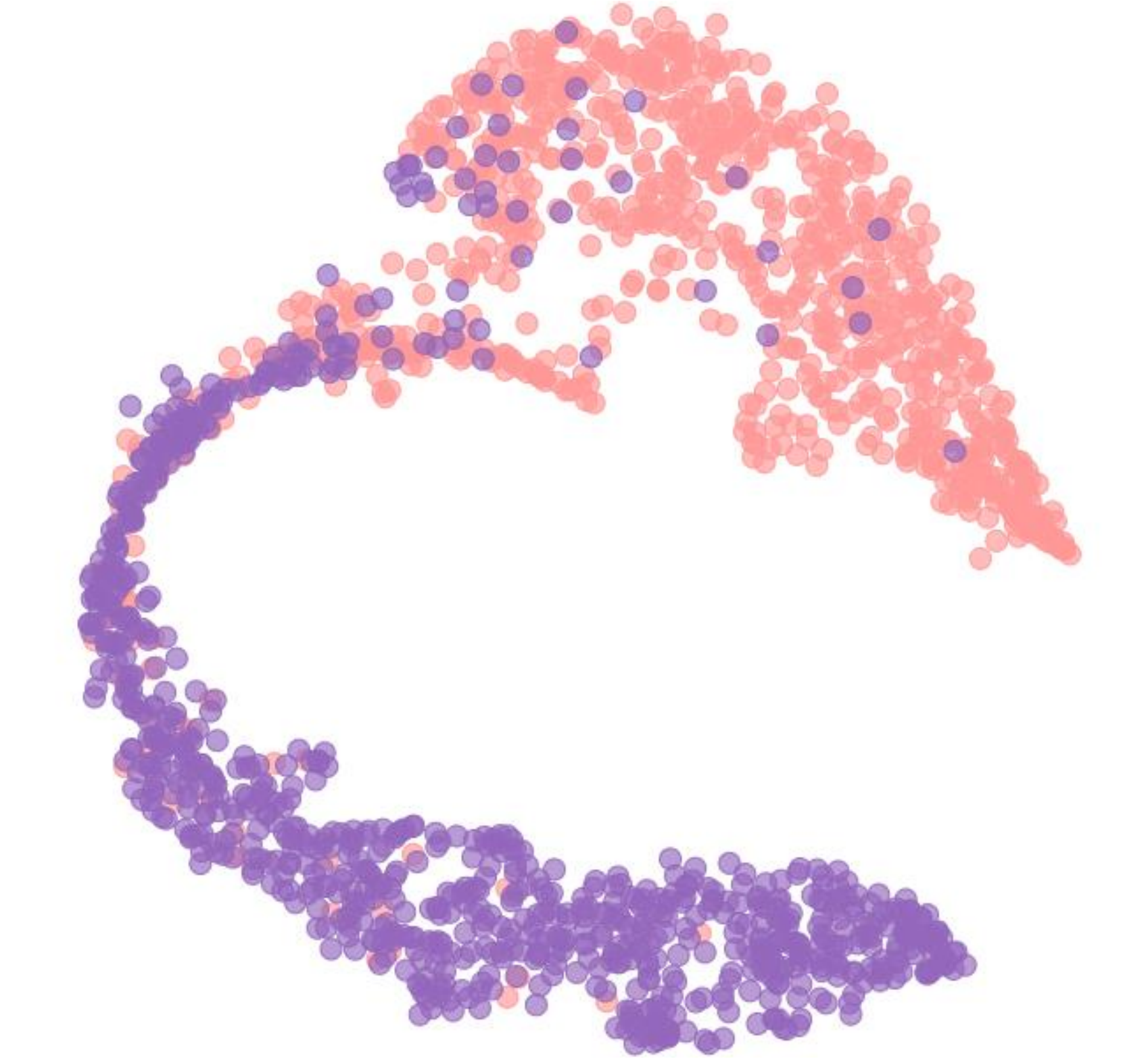}
        \caption{Multi-modal representations}
        \label{mm}
    \end{subfigure}
    \centering
    \begin{subfigure}{0.22\textwidth}
      \centering
        \includegraphics[height=0.92\linewidth]{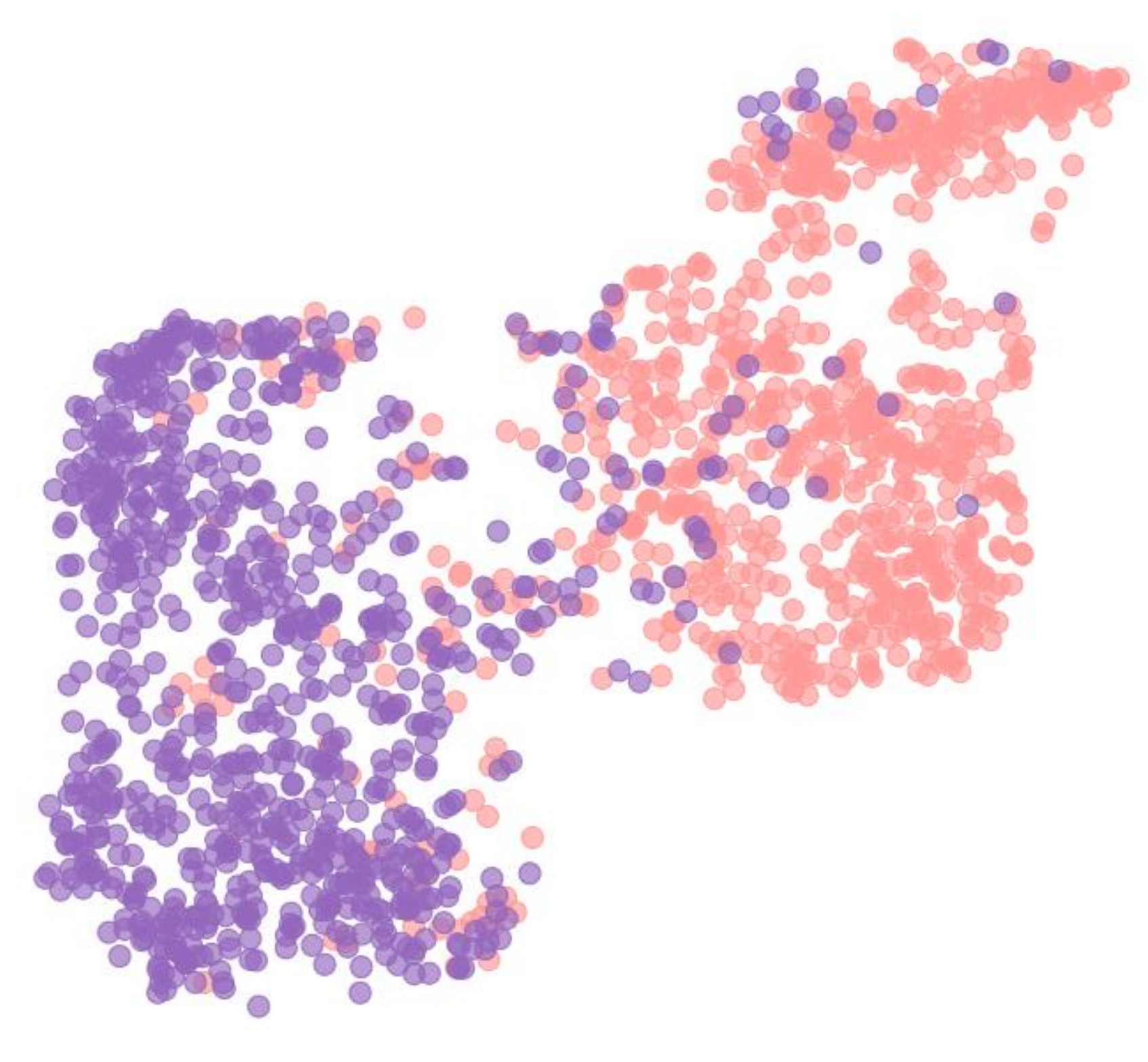}
        \caption{Completed representations}
        \label{comp}
    \end{subfigure}
            \caption{T-SNE visualizations of the different modal representations on the test dataset of Weibo.}
        \label{T-SNE}
    \centering
    \label{tSNE}
\end{figure}

\subsection{T-SNE Visualizations}


In Figure~\ref{tSNE}, we further analyze the separability of the different modal representations learned by MMFN and its variants on Weibo using t-SNE~\cite{van2008visualizing} visualizations. The visual, textual, multi-modal, and completed representations are obtained from MMFN w/o T, MMFN w/o I, MMFN w/o U, and completed MMFN, respectively. The dots of the same color indicate instances of the same label.

The visual representations learned by MMFN w/o T have some overlap with those of different labels, suggesting that the visual information alone is not sufficient for classification on Weibo. 
On the other hand, the textual representations learned by MMFN w/o I show a relatively better separability compared to the visual representations, indicating that the textual information can provide some discriminative information for the classification task.

The multi-modal representations learned by MMFN w/o U are more separable compared to the visual representations and are slightly better in separability to the textual representations. This indicates that the combination of textual and visual information can provide complementary information and improve the separability of the representations. However, due to the absence of uni-modal branches, MMFN w/o U isn't as effective as the completed model.
Finally, the representations learned by the completed MMFN show the best separability among all the variants, surpassing the separability of multi-modal representations. This suggests that the proposed method is effective in handling the usage of multi-modal representations to addressing the ambiguity problem, which significantly enhances the separability.

Overall, these results demonstrate not only the importance of combining both visual and textual information for fake news detection on Weibo, but also the effectiveness of the proposed fusion method in organizing multi-grained multi-modal representations.

\section{Conclusions}
In this paper, we present a novel multi-modal fake news detection method called MMFN, which uses BERT and Swin-T with a Transformer to obtain fine-grained multi-modal feature and uses CLIP to acquire coarse-grained multi-modal feature. In addition, we introduce uni-modal branches with CLIP similarity weighting to adaptively assist multi-modal classification. 
We conduct comprehensive experiments on two well-known datasets for multi-modal fake news detection.
The results show that MMFN outperforms many of the state-of-the-art methods.

\begin{acks}
This work is supported by the National Natural Science Foundation of China under Grants U20B2051, 62072114, U22B2047 and U1936214.
\end{acks}


\bibliographystyle{ACM-Reference-Format}
\bibliography{sample-sigconf}

\appendix

\end{document}